\documentclass[runningheads]{llncs}
\usepackage{graphicx}%
\usepackage{multirow}
\usepackage{amsmath,amssymb,amsfonts}%
\usepackage{mathrsfs}%
\usepackage[title]{appendix}%
\usepackage{xcolor}%
\usepackage{textcomp}%
\usepackage{manyfoot}%
\usepackage{booktabs}%
\usepackage{algorithm}%
\usepackage{algorithmicx}%
\usepackage{algpseudocode}%
\usepackage{listings}%
\usepackage{tabulary}
\usepackage{ragged2e}
\usepackage{tabularx}
\usepackage{subcaption}
\begin{document}
\title{NiNformer: A Network in Network Transformer with Token Mixing Generated Gating Function}
\titlerunning{NiNformer}
%
\author{Abdullah Nazhat Abdullah\inst{1}\orcidID{0000-0002-1757-0785} \and
Tarkan Aydin\inst{2}\orcidID{0000-0002-2018-405X} 
}
%
%
\institute{Bahcesehir University,Turkiye \\ \email{nazhat.abdullah@bahcesehir.edu.tr} \and Bahcesehir University,Turkiye \\ \email{tarkan.aydin@bau.edu.tr}}
\maketitle              

\begin{abstract}

The attention mechanism is the primary component of the transformer architecture; it has led to significant advancements in deep learning spanning many domains and covering multiple tasks. In computer vision, the attention mechanism was first incorporated in the Vision Transformer ViT, and then its usage has expanded into many tasks in the vision domain, such as classification, segmentation, object detection, and image generation. While the attention mechanism is very expressive and capable, it comes with the disadvantage of being computationally expensive and requiring datasets of considerable size for effective optimization. To address these shortcomings, many designs have been proposed in the literature to reduce the computational burden and alleviate the data size requirements. Examples of such attempts in the vision domain are the MLP-Mixer, the Conv-Mixer, the Perciver-IO, and many more attempts with different sets of advantages and disadvantages. This paper introduces a new computational block as an alternative to the standard ViT block. The newly proposed block reduces the computational requirements by replacing the normal attention layers with a Network in Network structure, therefore enhancing the static approach of the MLP-Mixer with a dynamic learning of element-wise gating function generated by a token mixing process. Extensive experimentation shows that the proposed design provides better performance than the baseline architectures on multiple datasets applied in the image classification task of the vision domain.

\end{abstract}
\newpage
\section{Introduction}
The advent of the transformer architecture \cite{1} and the introduction of the attention mechanism as its main computational component within the context of natural language processing (NLP) led to large advancements not only in language-related tasks but across all aspects related to the research and application of machine learning (ML). Transformers changed the landscape of NLP with the adoption of their architecture in designing highly successful and capable large language models (LLM) \cite{2} such as GPT \cite{3}, LLama \cite{4}, Falcon \cite{5}, and Mistral \cite{6}. The computer vision (CV) domain also experienced rapid adoption of transformer architectures. Vision-specific implementations such as ViT \cite{7}, MLP-Mixer \cite{8}, Conv-Mixer \cite{9}, and Swin Transformer \cite{10} were introduced, along with many application-oriented designs that utilize such architectures, such as Detection Transformer (DETR) \cite{11}, Perceiver-IO \cite{12}, Unified-IO \cite{13}, DINO \cite{14}, and Segment Anything Model (SAM) \cite{15}. In addition, efficiency-oriented implementations of the transformer architecture have been introduced, such as Linformer\cite{16}, FNets\cite{17}, Local-ViT\cite{18}, Max-ViT\cite{19}, and Nystromformer\cite{20}. These architectures introduce different types of trade-offs to increase the efficiency of models while reducing some of the technical aspects, such as the dynamic and full information mixing of the attention mechanism. In focus, a drawback of the MLP-Mixer design is that the information mixing processes are performed with static weight matrices, which limits the capabilities of the architecture in comparison to the traditional transformers that utilize the dynamic process of the scaled dot product attention mechanism with the softmax activation function. At the same time, the traditional transformer architecture has its own drawback of quadratic complexity in input size \cite{21}, which imposes a considerable cost in both training and inference when selecting the architecture. It is notable that in the literature there is a lack of a design that adopts the efficiency measures introduced in the MLP-Mixer model while also maintaining a dynamic information filtering mechanism, as with the traditional transformer design. In this paper, we introduce a newly formulated computational block that can be used as a core process in constructing transformer architectures that blends both efficient elementary operations and dynamic information filtering. The new proposal utilizes the MLP-Mixer token mixing to learn a generator of dynamic per input gating function that selectively filters the input representation tokens that are then passed to the per token MLP stage as in traditional transformers, which results in a block that contains two levels of processing \cite{22}, an inner and an outer, hence the chosen name for the proposal as a Network in Network Transformer, or (NiNformer). In this work, the newly proposed architecture was trained and its performance evaluated with respect to multiple baselines that represent different architectural directions and a variety of design choices. The comparison was conducted on three datasets, and the experimentation was performed in an equalized setting with the same computational resources to ensure a fair evaluation. From the experiments conducted, it was observed that the NiNformer architecture was the most performing, and the obtained results verified the validity and capability of the underlying assumptions employed in our proposed computational block.

\noindent The main contributions of our work are the following:
\begin{itemize}
\item A novel computational block that introduces a two-level Network in Network formulation to the design of transformer architecture.
\item An enhancement to static weight approaches of efficient Transformer designs by utilizing an MLP-Mixer as a subunit to generate a gating signal.
\item An introduction of a dynamic higher-level information processing that maintains a lower compute requirement than the scaled dot product attention mechanism.
\end{itemize}

\hfill

\section{Related Work}

\noindent The literature is rich with attempts to improve on the qualities and capabilities of the traditional transformer architecture design \cite{23},\cite{24},\cite{25},\cite{26}.These designs can be categorized into three main approaches:
\begin{itemize}
    \item Approximations of the attention mechanism
    \item Sparse and low-rank modifications of the attention mechanism
    \item Linear Alternatives to the attention mechanism
\end{itemize} 
This section is divided into three subsections following the categorization mentioned above.

\subsection{Approximations of Attention}

Guo et al. introduced Star Transformer \cite{27}, combining band attention and global attention. This formulation of the transformer has a global node on which a band attention of width 3 is applied. Also, a shared global node connects a pair of non-adjacent nodes, while adjacent nodes are connected to each other. Beltagy et al. introduced Longformer\cite{28}, which also uses a combination of band attention and internal global-node attention. Classification tokens are selected as global nodes. The architecture substitutes the band attention heads in the upper layers with dilated window attention, thus increasing the receptive field without increasing computation. Kitaev et al. introduced Reformer \cite{29} as a modified transformer that employs locality-sensitive hashing (LSH). The LSH is used to select the key and value pairs for each query, therefore allowing each token to attend to tokens that exist in the same hashing bucket. BigBird architecture by Zaheer et al. \cite{30} utilizes random attention to approximate full attention with a sparse encoder and sparse decoder, and it was shown by the analysis that this design can simulate any Turing Machine, explaining the capability of such architecture. Xiong et al. used the Nyström method to modify the transformer with the introduction of Nyströmformer \cite{20}. This design selects landmark nodes by the process of strided average pooling and then processes these selected queries and keys with an approximation to attention by the Nyström method. Katharopoulos et al. proposed the Linear Transformer \cite{31} with feature maps that target an approximation of the full scaled dot product attention with softmax activation function and showed comparable performance in empirical tests. Wang et al. introduced Linformer \cite{16}, showing an approximation to the attention mechanism by a low-rank matrix, thus lowering the computational requirement while maintaining comparable performance.Choromanski et al. proposed Performer \cite{32}, which uses random feature maps as an approximation to the traditional attention function. Tay et al. introduced the sparse Sinkhorn attention \cite{33}. This mechanism is essentially block-wise attention, but the keys are sorted block-wise, therefore learning the permutations.

\subsection{Sparse modifications of Attention}

\noindent Wang et al. introduced the Cascade Transformer \cite{34} By using a sliding window attention, the window size is exponentially increased when increasing the number of layers, leading to a reduction in complexity. Li et al. introduced the LogSparse Transformer \cite{35} that facilitates long-term dependency on time series analysis by using Eponym attention. Qiu et al. introduced BlockBERT \cite{36}, which uses block-wise attention to split the input sequence into non-overlapping blocks.  Dai et al. proposed the Transformer-XL\cite{37}. This design uses a recurrence between the windows that is segment-based. by storing the representations of the previous window and storing them in first-in, first-out memory (FIFO). After this step, the Transformer-XL applies attention to the sorted representations that have been stored in memory. Clustered Attention, proposed by Vyas et al. \cite{38} clusters the quires, then calculates the attention distributions for cluster centroids.Zhang et al. proposed PoolingFormer \cite{39}, which utilizes a two-level attention, a sliding window attention, and a compressed memory attention. The compressed memory module is used after first applying the sliding window attention, then applying a compressed memory module for the purpose of increasing the receptive field. Liu et al. proposed Memory Compressed Attention (MCA) \cite{40}, which complements local attention with strided convolution, thus reducing the number of keys and values. This allows the architecture to process much longer sequences compared to traditional transformers. Funnel Transformer \cite{41} was proposed by Dai et al. by employing a funnel-like encoder that has a gradual reduction of the hidden sequence length using pooling along the sequence dimension; the proper length is then restored with an up-sampling process.Max-ViT \cite{19} was introduced by Tu et al., which repeats the basic building block over multiple stages. The basic block consists of two aspects: blocked local attention and dilated global attention. Ho et al. proposed the Axial Transformer \cite{42}. This architecture computes a sequence of attention functions with each one applied along a single axis of the input, reducing the computational cost. Swin Transformer \cite{10} is an architecture proposed by Liu et al., and this design reduced the cost by splitting the image input into non-overlapping patches. These patches are then embedded as tokens for processing by Attention.

\subsection{Linear Alternatives to Attention}

\noindent FNets \cite{17} was introduced by Lee-Thorp et al., and it proposes an attention-free transformer architecture that substitutes the scaled dot product attention with softmax activation function. The Fourier sublayer applies a 2D DFT to the embedded input in two steps: one 1D DFT along the sequence dimension and another 1D DFT along the hidden dimension. gMLP \cite{43} was introduced by Liu et al., and this architecture is comprised of a series of blocks that are homogeneous in size and width. Each block layout is highly reminiscent of inverted bottlenecks. Another feature of this architecture compared to traditional transformers is that it does not require position embeddings. Local-ViT \cite{18} was introduced by Li et al. This architecture incorporates 2D depth-wise convolutions instead of the feed-forward network as in ViT. This design choice was inspired by the inverted residuals of MobileNets. Synthesizer \cite{44} was proposed by Tay et al. as an architecture that learns synthetic attention weights and does not rely on interactions between tokens. The results showed competitive performance in relation to other linear transformer designs. Transformer iN Transformer (TNT) \cite{45} was introduced by Han et al. This design treats the input images in a similar manner to a paragraph of text and divides them into several patches as “visual sentences” and then further divides them into sub-patches as “visual words”. With this hierarchical division, the architecture is divided into conventional transformer blocks for extracting features and attentions on the visual sentence level, and then a sub-transformer is introduced in order to extract the features of smaller visual words. De et al. proposed Hawk and Griffin models \cite{46}; these are hybrid models combining gated linear recurrences and local attention with good extrapolation capabilities.

\noindent The main shortcomings of the approaches previously attempted in the literature are the following:
\begin{itemize}
\item The use of static weight designs in order to increase efficiency results in loss of token to token interactions.
\item No attempt to recover dynamic token interactions in the previously introduced approaches.
\item Some approaches only modify the attention mechanism with kernel methods or approximations without a significant departure from the original design.
    
\end{itemize}

\section{Methodology}
The methodology section is divided into two subsections. In the first subsection, the baseline architectures used in the evaluation are outlined, followed by a second subsection where our proposed NiNformer architecture is described.

\subsection{Baselines}	
For an extensive comparative analysis of capability, our proposed architecture is contrasted to multiple baseline architectures that represent a variety of functional principles. The ViT follows the principles of a traditional NLP transformer, which represented the first iteration of designs that adopted such architecture. At its core, it relies on the scaled dot product attention with softmax activation function, and as with NLP-oriented transformers, the Vit also introduced the homogeneous layer structure.

\noindent Equations (1), (2), and (3) are the main equations for the ViT block.

\begin{equation}
\text{Attention(Q,K,V)} =\text{softmax}\left(\dfrac{QK^T}{\sqrt{d_k}}\right) V 
\end{equation}

\begin{equation}
\text{Y(X)} = 
\text{Attention}(\text{LayerNorm(X)})+ \text{X}
\end{equation}

\begin{equation}
\text{Z(Y)} = 
\text{MLP}(\text{LayerNorm(Y)})+\text{Y} 
\end{equation}

\noindent Procedure 1 overviews the ViT architecture.

\begin{algorithm}[H]
\caption{\textbf{Procedure 1 : }ViT}
\begin{algorithmic}

\State{\textbf{Input:}} Image $I$, number of classes $C$, patch size $ps$, embedding dimension $d_{model}$, number of Transformer blocks $B$, hidden dimension of MLP $d_{mlp}$, learning rate $\eta$

\State{\textbf{Output:}} Predicted class probabilities

\State{\textbf{Steps:}}

\hspace*{5mm}1. Divide $I$ into patches of size $ps \times ps$.

\hspace*{5mm}2. Flatten each patch and embed it into a $d_{model}$-dimensional vector using \hspace*{15mm}patch embedding layer.

\hspace*{5mm}3. Concatenate the embedded patches into a sequence $X$.

\hspace*{5mm}4. \textbf{for} $i = 1$ to $B$ \textbf{do:}

    \hspace*{15mm} Branch $X$ into residual and nonresidual paths.
    
    \hspace*{15mm} Normalize the nonresidual path and Apply Attention.
    
    \hspace*{15mm} Add the residual path.
    
    \hspace*{15mm} Branch Attention result into residual and nonresidual paths.
    
   \hspace*{15mm} Normalize the nonresidual path and Apply MLP block.
   
    \hspace*{15mm} Add the residual path.
   
\hspace*{5mm}5. Apply global average pooling to the output of the last Transformer block.

\hspace*{5mm}6. Use a fully connected layer with $C$ output units and softmax activation \hspace*{15mm}to obtain class probabilities.

\hspace*{5mm}7. Train the model by minimizing the loss between predicted and true labels \hspace*{15mm}using gradient descent with learning rate $\eta$.

\end{algorithmic}
\end{algorithm}

\noindent The MLP-Mixer adopts the homogeneous layer structure as with the ViT but introduces efficiency-oriented computational operations of mixing (interacting) the token representation with the application of MLP that are applied in two successive stages: first, an MLP mixing of per token representation, and second, a per position (channel) MLP mixing of representations in between the tokens.

\noindent Equations (4) and (5) are the main equation for the MLP-Mixer block.

\begin{equation}
\text{Y(X)} = \text{Transpose}(\text{MLP}(\text{Transpose}(\text{LayerNorm(X)})))
+\text{X} 
\end{equation}

\begin{equation}
\text{Z(Y)} = 
\text{MLP}(\text{LayerNorm(Y)})+\text{Y} 
\end{equation}

\noindent Procedure 2 overviews the MLP-Mixer architecture.

\begin{algorithm}[H]    
\caption{\textbf{Procedure 2 : }MLP-Mixer}
\begin{algorithmic}

\State{\textbf{Input:}} Image $I$, number of classes $C$, patch size $ps$, embedding dimension $d_{model}$, number of Transformer blocks $B$, hidden dimension of MLP $d_{mlp}$, learning rate $\eta$

\State{\textbf{Output:}} Predicted class probabilities

\State{\textbf{Steps:}}

\hspace*{5mm}1. Divide $I$ into patches of size $ps \times ps$.

\hspace*{5mm}2. Flatten each patch and embed it into a $d_{model}$-dimensional vector using \hspace*{15mm}patch embedding layer.

\hspace*{5mm}3. Concatenate the embedded patches into a sequence $X$.

\hspace*{5mm}4. \textbf{for} $i = 1$ to $B$ \textbf{do:}

    \hspace*{15mm} Branch $X$ into residual and nonresidual paths.
    
    \hspace*{15mm} Normalize the nonresidual path and Transpose.
    
    \hspace*{15mm} Apply MLP block.
    
    \hspace*{15mm} Transpose.
    
    \hspace*{15mm} Add the residual path.
    
    \hspace*{15mm} Branch result into residual and nonresidual paths.
    
   \hspace*{15mm} Normalize the nonresidual path and Apply MLP block.
   
    \hspace*{15mm} Add the residual path.
    
\hspace*{5mm}5. Apply global average pooling to the output of the last Transformer block.

\hspace*{5mm}6. Use a fully connected layer with $C$ output units and softmax activation \hspace*{15mm}to obtain class probabilities.

\hspace*{5mm}7. Train the model by minimizing the loss between predicted and true labels \hspace*{15mm}using gradient descent with learning rate $\eta$.

\end{algorithmic}
\end{algorithm}

\noindent The Local-ViT adopts a conservative design choice to introduce a more lightweight variant of the original ViT by replacing the per-token MLP layer in the ViT block with convolutions.

\noindent Equations (6) and (7) are the main equations for the Local-ViT block.

\begin{equation}
\text{Y}(X) = 
\text{Attention}(\text{LayerNorm(X)})+ \text{X}
\end{equation}

\begin{equation}
\text{Z}(Y) = 
\text{CONV}(\text{LayerNorm(Y)})+\text{Y} 
\end{equation}

\noindent Procedure 3 overviews the Local-ViT architecture.

\begin{algorithm}[H]
\caption{\textbf{Procedure 3 : }Local-ViT}
\begin{algorithmic}

\State{\textbf{Input:}} Image $I$, number of classes $C$, patch size $ps$, embedding dimension $d_{model}$, number of Transformer blocks $B$, hidden dimension of MLP $d_{mlp}$, learning rate $\eta$

\State{\textbf{Output:}} Predicted class probabilities

\State{\textbf{Steps:}}

\hspace*{5mm}1. Divide $I$ into patches of size $ps \times ps$.

\hspace*{5mm}2. Flatten each patch and embed it into a $d_{model}$-dimensional vector using \hspace*{15mm}patch embedding layer.

\hspace*{5mm}3. Concatenate the embedded patches into a sequence $X$.

\hspace*{5mm}4. \textbf{for} $i = 1$ to $B$ \textbf{do:}

    \hspace*{15mm} Branch $X$ into residual and nonresidual paths.
    
    \hspace*{15mm} Normalize the nonresidual path and Apply Attention.
    
    \hspace*{15mm} Add the residual path.
    
    \hspace*{15mm} Branch Attention result into residual and nonresidual paths.
    
   \hspace*{15mm} Normalize the nonresidual path and Apply CONV block.
   
    \hspace*{15mm} Add the residual path.
    
\hspace*{5mm}5. Apply global average pooling to the output of the last Transformer block.

\hspace*{5mm}6. Use a fully connected layer with $C$ output units and softmax activation \hspace*{15mm}to obtain class probabilities.

\hspace*{5mm}7. Train the model by minimizing the loss between predicted and true labels \hspace*{15mm}using gradient descent with learning rate $\eta$.

\end{algorithmic}
\end{algorithm}

\subsection{Proposed Architecture}
The proposed computational block of this paper is comprised of two levels: an outer network that resembles a transformer block by including a token-wise MLP, which provides the design with an optimization-driven token mapping capability. The token-wise MLP of the outer network is preceded in the proposed block by a substitute for the attention mechanism, which has a gating function process on the outer network level that extends the concept of gated linear unit (GLU) \cite{47} by employing a Network in Network structure. In the proposed gating unit, the gating signal is generated by a sub-unit in the inner network, where the inner sub-unit uses a token-mixing architecture of the MLP-Mixer. The proposed design significantly differs from TNT architecture \cite{45} in that the two levels in our proposal are different in form and function, and both inner and outer levels apply their transformations to the input context as a whole, while the TNT architecture has two levels of the same traditional attention mechanism that are applied on two separate scales, the visual word scale and the visual sentence scale within the input context. Such distinction of scales omits processing of the global correlations that may exist between parts of the context in the case of TNT, and our design utilizes the full context on both of its two levels to capture the global correlations of the input.In addition, the newly introduced gating mechanism has the advantage of using the non-dynamic, fixed-weight MLP-Mixer as an inner sub-unit to learn the interdependencies from the input representation, which is then used by the outer level as a dynamic gating signal that functions on an input by input basis to scale the values of its linearly projected representation, thus facilitating further information processing by the outer level MLPs without the use of the scaled dot product attention employed in generic transformer architectures. The two levels of our proposal rely on element-wise operations, as both the gating operation and the internal MLP-Mixer are based on linear complexity element-wise multiplications, making our proposal of $O(n)$ complexity.

\noindent Equations (8), (9) and (10) describe the operation of the proposed block.

\begin{equation}
\text{Gating(I)} = 
(\text{MLPMixer(I)}) * \text{Linear(I)}
\end{equation}

\begin{equation}
\text{Y(X)} = 
\text{Gating}(\text{LayerNorm(X)})+ \text{X}
\end{equation}

\begin{equation}
\text{Z(Y)} = 
\text{MLP}(\text{LayerNorm(Y)})+\text{Y} 
\end{equation}

\noindent Procedure 4 overviews our proposed NiNformer architecture.

\begin{algorithm}[H]
\caption{\textbf{Procedure 4 : }NiNformer}
\begin{algorithmic}
    
\State{\textbf{Input:}} Image $I$, number of classes $C$, patch size $ps$, embedding dimension $d_{model}$, number of Transformer blocks $B$, hidden dimension of MLP $d_{mlp}$, learning rate $\eta$

\State{\textbf{Output:}} Predicted class probabilities

\State{\textbf{Steps:}}

\hspace*{5mm}1. Divide $I$ into patches of size $ps \times ps$.

\hspace*{5mm}2. Flatten each patch and embed it into a $d_{model}$-dimensional vector using \hspace*{15mm}patch embedding layer.

\hspace*{5mm}3. Concatenate the embedded patches into a sequence $X$.

\hspace*{5mm}4. \textbf{for} $i = 1$ to $B$ \textbf{do:}

    \hspace*{15mm} Branch $X$ into residual and nonresidual paths.
    
    \hspace*{15mm} Normalize the nonresidual path
    
    \hspace*{15mm} Generate the gating signal by the application of the MLP-Mixer \hspace*{24mm}sub-unit on the nonresidual path.

    \hspace*{15mm} Apply the Gating by multiplying the liearly projected nonresidual \hspace*{24mm}path with the MLP-Mixer sub-unit output.
    
    \hspace*{15mm} Add the residual path.
    
    \hspace*{15mm} Branch Gating result into residual and nonresidual paths.
    
   \hspace*{15mm} Normalize the nonresidual path and Apply MLP block.
   
    \hspace*{15mm} Add the residual path.
    
\hspace*{5mm}5. Apply global average pooling to the output of the last Transformer block.

\hspace*{5mm}6. Use a fully connected layer with $C$ output units and softmax activation \hspace*{15mm}to obtain class probabilities.

\hspace*{5mm}7. Train the model by minimizing the loss between predicted and true labels \hspace*{15mm}using gradient descent with learning rate $\eta$.

\end{algorithmic}
\end{algorithm}

\begin{figure}[H]
    \centering
    \begin{subfigure}[t]{0.5\textwidth}
        \centering
        \includegraphics[width=\textwidth,height=2.0in]{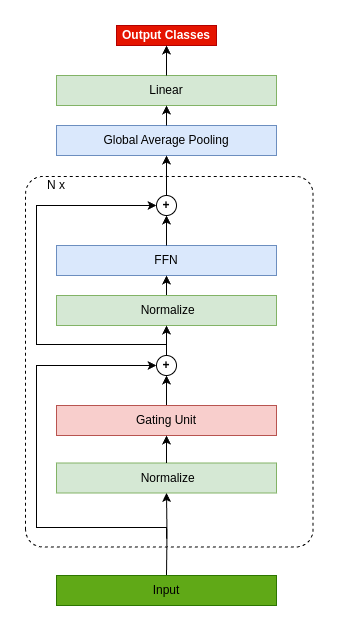}
        \caption{NiNformer architecture}
    \end{subfigure}%
    ~ 
    \begin{subfigure}[t]{0.5\textwidth}
        \centering
        \includegraphics[width=\textwidth,height=2.0in]{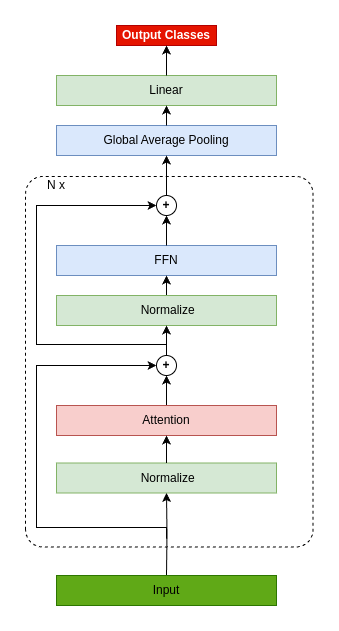}
        \caption{ViT architecture}
    \end{subfigure}     
    ~    
    \begin{subfigure}[t]{0.5\textwidth}
        \centering
        \includegraphics[width=\textwidth,height=2.0in]{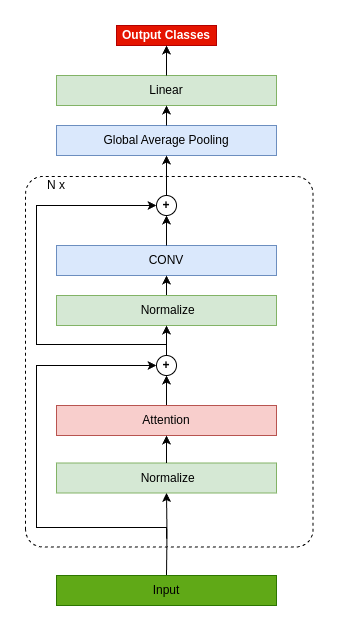}
        \caption{Local-ViT architecture}
    \end{subfigure}

    ~ 
    \begin{subfigure}[t]{0.5\textwidth}
        \centering
        \includegraphics[width=\textwidth,height=2.0in]{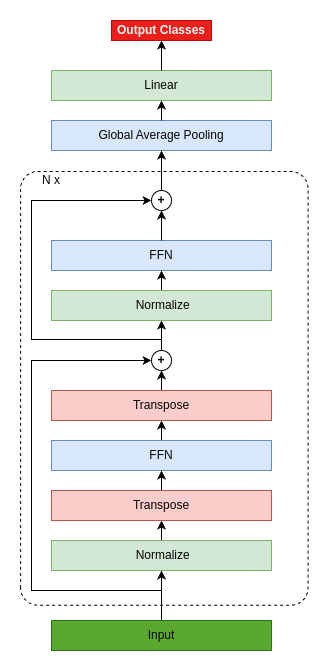}
        \caption{MLP-Mixer architecture}
    \end{subfigure}%
   ~
    \caption{A diagrammatic comparison of NiNformer architecture with ViT, MLP-Mixer and Local-ViT.}
    \label{architectures}
\end{figure}

\begin{figure}[H]
  \centering
  \begin{tabular}{@{}c@{}}
   \includegraphics[width=\linewidth,height=3.5in]{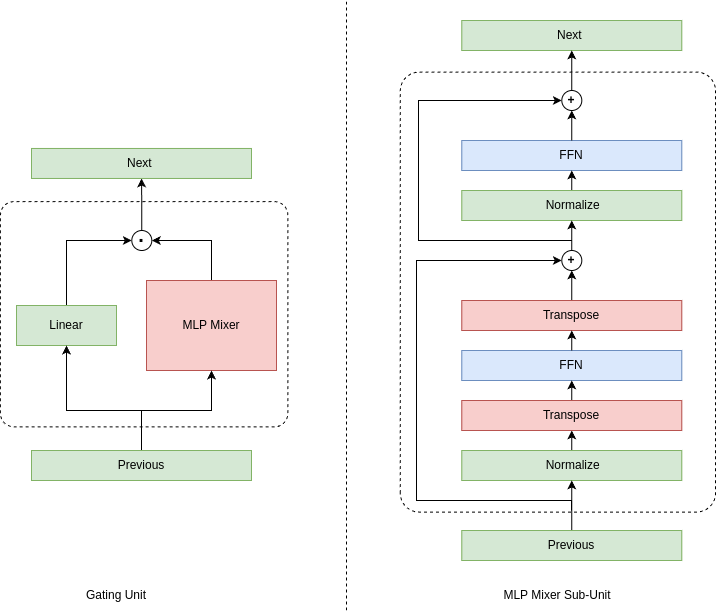} \\[\abovecaptionskip]
    \small (a) NiNformer gating-unit and Mixer sub-unit
  \end{tabular}

   \begin{tabular}{@{}c@{}}
   \includegraphics[width=3.5in,height=3.5in]{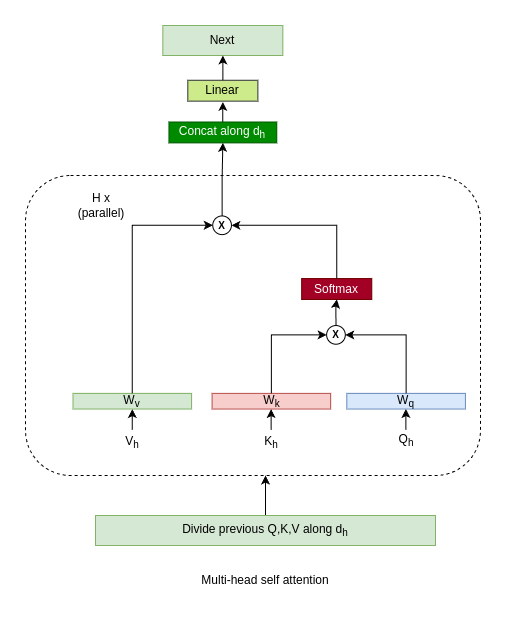} \\[\abovecaptionskip]
    \small (b) Multi-head self attention
  \end{tabular}

  \caption{A diagrammatic comparison of NiNformer mechanism with the attention mechanism.}\label{Blocks}
\end{figure}

\noindent Fig. \ref{architectures} shows the NiNformer overall architecture in comparison to the Vit, MLP-Mixer and Local-ViT architectures, while Fig. \ref{Blocks} compares the proposed NiNformer mechanism with the attention mechanism of ViT.

\noindent Table 1 illustrates the advantages and disadvantages of our proposed design in comparison to the baseline architectures.

\begin{table}[htbp]

\caption{Advantages and disadvantages of baseline architectures and proposed architecture.}
\begin{tabularx}{5in}{XXXX}\toprule

 Architecture &Advantages  &Disadvantages  &Compute requirement (time/memory)    \\ \cmidrule{1-4}
 ViT & global token to token interaction   & attention is quadratic  & high compute requirement   \\ \cmidrule{1-4}
 Local-ViT & convolution lowers compute requirement  & inductive bias of CNN  & moderate compute requirement   \\  \cmidrule{1-4}
 MLP-Mixer& use of MLPs only   & loss of dynamic token interaction & low compute requirement \\ \cmidrule{1-4}
 \textbf{NiNformer (ours)} & the gating function ensures dynamic token interaction, while the use of the MLP-Mixer sub-unit lowers the compute requirement   & token to token processing is not fully global  &low compute requirement  \\ \bottomrule
\end{tabularx}

\end{table}

\noindent The Network in Network formulation proposed in this work solves the loss of dynamic token interaction that the MLP-Mixer approach suffers from by incorporating it as a learned gating signal generation sub-unit. Our design maintains the advantage of linear complexity provided by element-wise multiplication, gaining the advantage of low computational requirements in comparison to the traditional ViT transformer and avoiding the inductive bias-introducing mechanisms such as the convolutions utilized by the Local-ViT approach.
 
\section{Results}
For the purposes of experimental evaluation, three data sets have been selected as follows:
\begin{enumerate}
\item The CIFAR-10 \cite{48} dataset consists of 60000 color images in 32 by 32 resolution provided for 10 classes, with 6000 images per class. There are 50000 training images and 10000 test images.
\item The CIFAR-100 \cite{48}dataset consists of 60000 color images in 32 by 32 resolution; the number of classes is 100, resulting in 600 images per class. Similar to CIFAR-10, there are 50000 training images and 10000 test images. \item The MNIST \cite{49} dataset consists of 70,000 grayscale images in 28 by 28 resolution. The number of classes is 10, as it is a dataset of handwritten numerical digits. There are 60000 training images and 10000 test images.
\end{enumerate}
The utilized software tools are as follows:\begin{enumerate}
\item Python programming language of version 3.9.
\item Pytorch framework of version 1.13.
\item NVIDIA CUDA toolkit, of version 11.6.2.\end{enumerate}

\noindent The available hardware system is specified as follows:\begin{enumerate}
\item Intel i9-9900k CPU.
\item 32 Gigabytes of system RAM.
\item Nvidia RTX 2080ti GPU with 12 Gigabytes of VRAM.
\item UBUNTU 20 LTS operating system.\end{enumerate}
The implementation details of the selected transformer architectures in this work are as follows:
\begin{enumerate}
\item For the ViT architecture, the chosen patch size was 4 with a token dimension of 256, and the number of layers chosen was 4 with 4 attention heads and an MLP dimension of 512.
\item For the MLP-Mixer architecture, the chosen patch size was 4 with a token dimension of 256, and the number of layers chosen was 4 with a token-wise MLP dimension of 512 and a channel-wise MLP dimension of 512.
\item For the Local-ViT architecture, the chosen patch size was 4 with a token dimension of 256, the number of layers chosen was 4, 4 attention heads were selected, and the chosen channel dimension of the feedforward part was 512.
\item For the NiNformer architecture, the chosen patch size was 4, the number of layers chosen was 4, the token dimension selected was 256, and the MLP dimension was 512 in the outer network. The inner sub-unit was designed with a token-wise MLP dimension of 512 and a channel-wise MLP dimension of 512.
\end{enumerate} 

\noindent All models were fitted with a training loop comprised of 100 epochs with a batch size of 128. All experiments adopted the recommended learning rate for the Adam optimizer of 0.001 \cite{50}, other hyper-parameters such as patch size and token dimension were chosen so that it saturates the hardware capacity provided by the available computer system.
\newpage 
\noindent Table 2 illustrates the obtained results after performing the experimentation on MNIST, CIFAR-10 and CIFAR-100 datasets applied to the baseline architectures and NiNformer architecture.

\begin{table}[htbp]
\caption{Experimental test accuracy in percentages (\%) obtained on the utilized dataset.}
\begin{tabular}{lrrr}
\toprule
\multirow{2}{*}{Models} & \multicolumn{3}{l}{Data sets}                                                                   \\ \cmidrule(l){2-4}
                        & \multicolumn{1}{p{30mm}}{MNIST}          & \multicolumn{1}{p{30mm}}{CIFAR-10}        & \multicolumn{1}{p{30mm}}{CIFAR-100} \\ \cmidrule{1-4}
ViT                     & \multicolumn{1}{l}{97.12}          & \multicolumn{1}{l}{65.74}          &\multicolumn{1}{l} {34.87}                         \\ \cmidrule{1-4}
MlpMixer                & \multicolumn{1}{l}{97.73}          & \multicolumn{1}{l}{70.12}          & \multicolumn{1}{l}{39.16}                         \\ \cmidrule{1-4}
LocalViT                & \multicolumn{1}{l}{97.79}          & \multicolumn{1}{l}{77.71}          & \multicolumn{1}{l}{41.61}                         \\ \cmidrule{1-4}
NiNformer \textbf{(ours)}        & \multicolumn{1}{l}{\textbf{98.61}} & \multicolumn{1}{l}{\textbf{81.59}} & \multicolumn{1}{l}{\textbf{53.78}}                \\ \bottomrule
\end{tabular}
\end{table}

\noindent The performance of deep learning models is highly dependent on the low-level hardware details and software optimizations \cite{51}; the timing of execution shows significant sensitivity to the interactions between micro-architectural and execution characteristics such as caches and RAM configurations.We have performed per-sample inference time measurements on the selected baseline architectures and the proposed architecture of this work conducted as relative performance measures in relation to the hardware system available for the purposes of this work.
\noindent Table 3 illustrates the obtained inference time results after performing the experimentation on MNIST, CIFAR-10 and CIFAR-100 datasets applied to the baseline architectures and NiNformer architecture.

\begin{table}[htbp]

\caption{Experimental per-sample inference time measured in nano-seconds obtained on the utilized dataset.}
\begin{tabular}{lrrr}
\toprule
\multirow{2}{*}{Models} & \multicolumn{3}{l}{Data sets}                                                                   \\ \cmidrule(l){2-4}
                        & \multicolumn{1}{p{30mm}}{MNIST}          & \multicolumn{1}{p{30mm}}{CIFAR-10}        & \multicolumn{1}{p{30mm}}{CIFAR-100} \\ \cmidrule{1-4}
ViT                     & \multicolumn{1}{l}{141.62}          & \multicolumn{1}{l}{142.64}          &\multicolumn{1}{l} {141.07}                         \\ \cmidrule{1-4}
MlpMixer                & \multicolumn{1}{l}{132.68}          & \multicolumn{1}{l}{103.53}          & \multicolumn{1}{l}{104.24}                         \\ \cmidrule{1-4}
LocalViT                & \multicolumn{1}{l}{139.65}          & \multicolumn{1}{l}{127.76}          & \multicolumn{1}{l}{115.32}                         \\ \cmidrule{1-4}
NiNformer \textbf{(ours)}        & \multicolumn{1}{l}{135.00} & \multicolumn{1}{l}{104.64} & \multicolumn{1}{l}{105.37}                \\ \bottomrule
\end{tabular}

\end{table}

\noindent The obtained measurements are in support of the design goals of our proposals, as the inference time of our work adds a low inference time cost on the MLP-Mixer, which is used as a subunit within our work.Taking the measurements on CIFAR-10 as a reference, the execution time cost is only an additional 1\%, while the improvement in accuracy over the MLP-Mixer is significant at 16\%. Extending the comparison to the other baselines, our proposal shows a wide gain in accuracy of 24\% and 16\% for ViT and Local-ViT respectively, while also gaining a significant improvement margin in inference time measurements of 36\% and 22\% for ViT and Local-ViT, respectively.The results are in high accordance with the hypothesis introduced in this work of formulating a novel computational block that enhances the capacity and capability of linear alternatives to the attention mechanism while maintaining the properties of efficiency and fast execution margins over the traditional formulation of Transformers.

\begin{figure}[htbp]
    \centering
    \begin{subfigure}[t]{0.5\textwidth}
        \centering
        \includegraphics[width=\textwidth,height=1.3in]{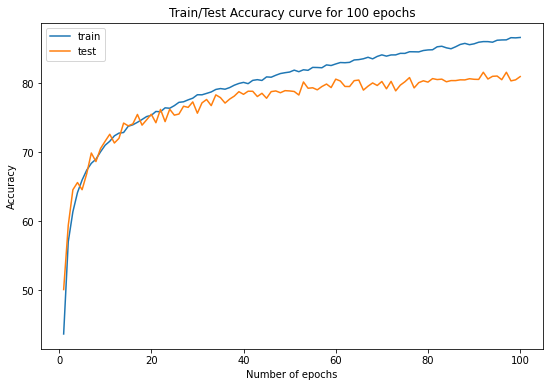}
        \caption{CIFAR-10 accuracy curve}
    \end{subfigure}%
    ~ 
    \begin{subfigure}[t]{0.5\textwidth}
        \centering
        \includegraphics[width=\textwidth,height=1.3in]{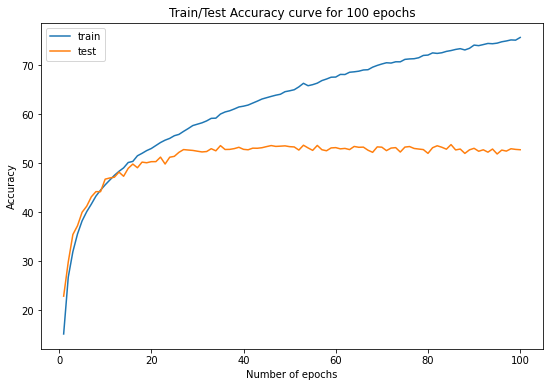}
        \caption{CIFAR-100 accuracy curve}
    \end{subfigure}
      ~ 
    \begin{subfigure}[t]{0.5\textwidth}
        \centering
        \includegraphics[width=\textwidth,height=1.3in]{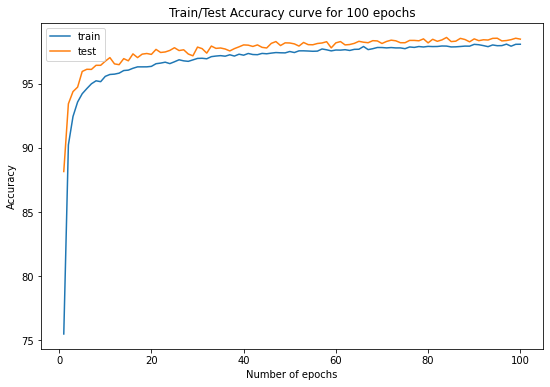}
        \caption{MNIST accuracy curve}
    \end{subfigure}
    \caption{An illustration of the accuracy curves for NiNformer architecture.}

    \label{accuracy}
\end{figure}

\begin{figure}[htbp]
    \centering
    \begin{subfigure}[t]{0.5\textwidth}
        \centering
        \includegraphics[width=\textwidth,height=1.3in]{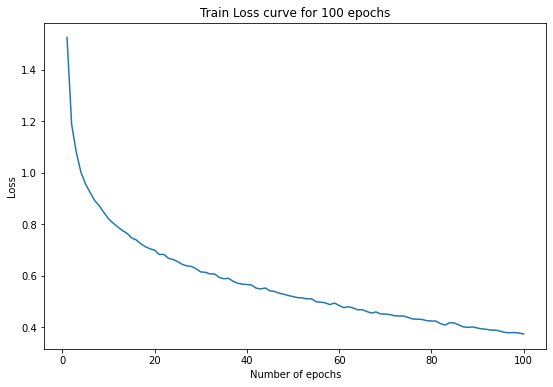}
        \caption{CIFAR-10 loss curve}
    \end{subfigure}%
    ~ 
    \begin{subfigure}[t]{0.5\textwidth}
        \centering
        \includegraphics[width=\textwidth,height=1.3in]{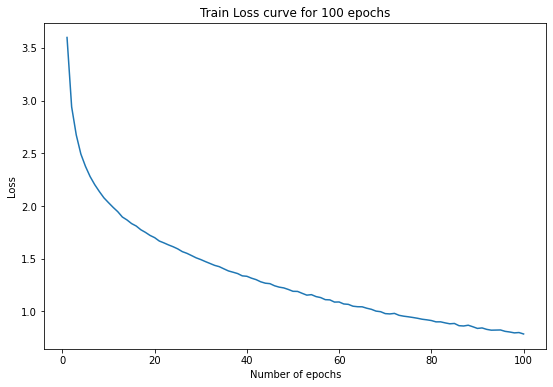}
        \caption{CIFAR-100 loss curve}
    \end{subfigure}
      ~ 
    \begin{subfigure}[t]{0.5\textwidth}
        \centering
        \includegraphics[width=\textwidth,height=1.3in]{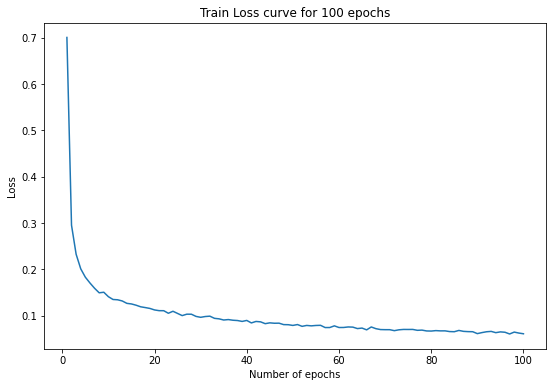}
        \caption{MNIST loss curve}
    \end{subfigure}
    \caption{An illustration of the loss curves for NiNformer architecture.}

    \label{loss}
\end{figure}

\noindent Fig. \ref{accuracy} and Fig. \ref{loss} show the accuracy and loss curves obtained on NiNformer for the CIFAR-10, CIFAR-100, and MNIST datasets.

\newpage

\newpage
\section{Conclusion}
This work introduced a newly designed Network in Network block that substitutes the attention block traditionally utilized in designing transformer architectures. The proposed efficient and highly performing block extends the token mixing approach presented in the MLP-Mixer to function as a gating signal generator and takes advantage of the gating mechanism to introduce dynamic token processing. The new mechanism of our proposal presents an enhancement of the static weight approach of the MLP-Mixer by utilizing its layers as a sub-unit network incorporated within a gating function of an outer network formulation. The experimental results show that our proposed block significantly outperforms the baseline architectures, offering noticeable improvements on the selected baselines, specifically showing a great enhancement of accuracy compared to the standalone MLP-Mixer architecture that acts as a sub-unit, validating the assumptions of the proposal introduced in this work positing that a two-level Network in Network organization of the main computational block and employing a dynamic gating of the upstream representation results in a significant enhancement and circumvents the shortcoming of the static weight approach of the standalone MLP-Mixer while still providing more simplicity of operations in contrast to the vanilla ViT transformer architecture. Future directions of this work are to investigate a multitude of sub-unit network selections, aiming for further enhancements and capabilities.

\end{document}